\documentclass{article}

\usepackage{arxiv}

\usepackage[utf8]{inputenc} 
\usepackage[T1]{fontenc}    
\usepackage{hyperref}       
\usepackage{url}            
\usepackage{booktabs}       
\usepackage{amsfonts}       
\usepackage{nicefrac}       
\usepackage{microtype}      
\usepackage{lipsum}		
\usepackage{graphicx}
\usepackage{natbib}
\usepackage{doi}
\usepackage{url,hyperref,lineno,microtype,subcaption}
\usepackage[onehalfspacing]{setspace}
\usepackage{bm}
\usepackage{amsmath}
\usepackage{multirow}

\title{Learning suction graspability considering grasp quality and robot reachability for bin-picking}

\author{ {}\\
{Ping Jiang$^{*}$, Junji Oaki, Yoshiyuki Ishihara, Junichiro Ooga, Haifeng Han,}\\
{Atsushi Sugahara, Seiji Tokura, Haruna Eto, Kazuma Komoda, and Akihito Ogawa}\\
Corporate Research \& Development Center\\
Toshiba Corporation\\
1, Komukai-Toshiba-cho, Saiwai-ku, Kawasaki 212-8582, Japan.\\
	\texttt{ping2.jiang@toshiba.co.jp}\\
}

\renewcommand{\shorttitle}{\textit{arXiv} Template}

\hypersetup{
pdftitle={A template for the arxiv style},
pdfsubject={q-bio.NC, q-bio.QM},
pdfauthor={Ping Jiang},
pdfkeywords={bin picking, grasp planning, suction grasp, graspability, deep learning},
}

\begin{document}
\maketitle

\begin{abstract}
Deep learning has been widely used for inferring robust grasps. Although human-labeled RGB-D datasets were initially used to learn grasp configurations, preparation of this kind of large dataset is expensive. To address this problem, images were generated by a physical simulator, and a physically inspired model (e.g., a contact model between a suction vacuum cup and object) was used as a grasp quality evaluation metric to annotate the synthesized images. However, this kind of contact model is complicated and requires parameter identification by experiments to ensure real world performance. In addition, previous studies have not considered manipulator reachability such as when a grasp configuration with high grasp quality is unable to reach the target due to collisions or the physical limitations of the robot. In this study, we propose an intuitive geometric analytic-based grasp quality evaluation metric. We further incorporate a reachability evaluation metric. We annotate the pixel-wise grasp quality and reachability by the proposed evaluation metric on synthesized images in a simulator to train an auto-encoder--decoder called suction graspability U-Net++ (SG-U-Net++). Experiment results show that our intuitive grasp quality evaluation metric is competitive with a physically-inspired metric. Learning the reachability helps to reduce motion planning computation time by removing obviously unreachable candidates. The system achieves an overall picking speed of 560 PPH (pieces per hour).

\end{abstract}


\section{Introduction}

In recent years, growth in retail e-commerce (electronic-commerce) business has led to high demand for warehouse automation by robots \citep{1}. Although the Amazon picking challenge \citep{2} has advanced the automation of the pick-and-place task, which is a common task in warehouses, picking objects from a cluttered scene remains a challenge.

The key to automation of pick-and-place is to find the grasp point where the robot can approach via a collision free path and then stably grasp the target object. Grasp point detection methods can be broadly divided into analytical and data-driven methods. Analytical methods \citep{3,4} require modeling the interaction between the object and the hand, and have a high computation cost \citep{5}. For those reasons, data-driven methods are preferred for bin picking.

Many previous studies have used supervised deep learning, which is one of the most widely used data-driven method, to predict only grasp point configuration (e.g., location, orientation, and open width) without considering the grasp quality. Given an RGB-D image, the grasp configuration for a jaw gripper \citep{6,7,8} or a vacuum gripper \citep{9,10} can be directly predicted using a deep convolutional neural network (DCNN). Learning was extended from points to regions by Domae et al. \citep{11, 12}, who proposed a convolution-based method in which the hand shape mask is convolved with the depth mask to obtain the region of the grasp points. Matsumura et al. \citep{13} later learned the peak among all regions for different hand orientations to detect a grasp point capable of avoiding multiple objects. 

However, in addition to the grasp configuration, the grasp quality is also important for a robot to select the optimal grasp point for bin picking. The grasp quality indicates the graspable probability by considering factors such as surface properties. For example, for suction grasping, although an object with a complicated shape may have multiple grasp points, the grasp points located on flat surfaces need to be given a higher selection priority because they have higher grasp quality (easier for suction by vacuum cup) than do curved surfaces. Zeng et al. \citep{14} empirically labeled the grasp quality in the RGB-D images of the Amazon picking challenge object set. They proposed a multi-modal DCNN for learning grasp quality maps (pixel-wise grasp quality corresponding to an RGB-D image) for jaw and vacuum grippers. However, preparing a dataset by manual labeling is time consuming and so the dataset was synthesized in a simulator to reduce the time cost. Dex-Net \citep{15,16} evaluated the grasp quality by a physical model and generated a large dataset by simulation. They used the synthesized dataset to train a grasp quality conventional neural network (GQ-CNN) to estimate the success probability of the grasp point. However, defining a precise physical model for contact between gripper and object is difficult. Further, the parameters of the model needed to be identified experimentally to reproduce the salient kinematics and dynamics features of a real robot hand (e.g., the deformation and suction force of a vacuum cup). 

Unlike Dex-Net, this study proposes an intuitive suction grasp quality analytic metric based on point clouds without the need for modeling complicated contact dynamics. Further, we incorporate a robot reachability metric to evaluate the suction graspability from the viewpoint of the manipulator. Previous studies have evaluated grasp quality only in terms of grasp quality for the hand only. However, it is possible that although a grasp point has high grasp quality, the manipulator is not able to move to that point. It is also possible for an object to have multiple grasp points with same the level of graspability but varying amounts of time needed for the manipulator to approach due to differences in the goal pose and surrounding collision objects. Bin picking efficiency can therefore be improved by incorporating a reachability evaluation metric. We label suction graspability by the proposed grasp quality and reachability metric and generate a dataset by physical simulator. An auto-encoder is trained to predict the suction graspability given the depth image input and a graspability clustering and ranking algorithm is designed to propose the optimal grasp point.  

Our primary contributions include:
1) Proposal of an intuitive grasp quality evaluation metric without complicated physical modeling. 
2) Proposal of a reachability evaluation metric for labeling suction grapability in addition to grasp quality. 
3) Performance of a comparison experiment between the proposed intuitive grasp quality evaluation metric and a physically-inspired one (Dex-Net). 
4) Performance of  an experiment to investigate the effect of learning reachability.

\section{Related works}

\subsection{Pixel-wise graspability learning}
In early studies, deep neural networks were used to directly predict the candidate grasp configurations without considering the grasp quality \citep{23,24,25}. However, since there can be multiple grasp candidates for an object that has a complicated shape or multiple objects in a cluttered scene, learning graspablity is required for the planner to find the optimal grasp among the candidates. 

Pixel-wise graspablity learning uses RGB-D or depth-only images to infer the grasp success probability at each pixel. One representative work is \citep{14} by Zeng et al., who used a manually labeled dataset to train fully convolutional networks (FCNs) for predicting pixel-wise grasp quality (affordance) maps of four pre-defined grasping primitives. Liu et al. \citep{55} performed active exploration by pushing objects to find good grasp affordable maps predicted by Zeng's FCNs. Recently, Utomo et al. \citep{43} modified the architecture of Zeng's FCNs to improve the inference precision and speed. Based on Zeng's concept, Hasegawa et al. \citep{54} incorporated a primitive template matching module, making the system adaptive to changes in grasping primitives. Zeng et al. also applied the concept of pixel-wise affordance learning to other manipulation tasks such as picking by synergistic coordination of push and grasp motions \citep{44}, and picking and throwing \citep{45}. However, preparing huge amounts of RGB-D images and manually labeling the grasp quality requires a large amount of effort.

Faced with dataset generation cost of RGB-D based graspability learning, researchers started to use depth-image-only based learning. The merits of using depth images is that they are easier to synthesize and annotate in a physical simulator compared with RGB images. Morrison et al. \citep{46} proposed a generative grasping convolutional neural network (GG-CNN) to rapidly predict pixel-wise grasp quality. Based on a similar concept of grasp quality learning, the U-Grasping fully convolutional neural network (UGNet) \citep{52}, Generative Residual Convolutional Neural Network (GRConvNet) \citep{50}, and Generative
Inception Neural Network (GI-NNet) \citep{53} were later proposed and were reported to achieve higher accuracy than GG-CNN. Le et al. \citep{51} extended GG-CNN to be capable of predicting the grasp quality of deformable objects by incorporating stiffness information. Morrison et al. \citep{49} also applied GG-CNN to a multi-view picking controller to avoid bad grasp poses caused by occlusion and collision. However, the grasp quality dataset of GG-CNN was generated by creating masks of the center third of each grasping rectangle of the Cornell Grasping dataset \citep{47} and Jacquard dataset \citep{48}. This annotation method did not deeply analyze the interaction between hand and object, which is expected to lead to insufficient representation of grasp robustness. 

To improve the robustness of grasp quality annotation, a physically-inspired contact force model was designed to label pixel-wise grasp quality. Mahler et al. \citep{15,16} designed a quasi-static spring model for the contact force between the vacuum cup and the object. Based on the designed compliant contact model, they assessed the grasp quality in terms of grasp robustness in a physical simulator. They further proposed GQ-CNN to learn the grasp quality and used a sampling-based method to propose an optimal grasp in the inference phase, and also extended their work by proposing a fully convolutional GQ-CNN \citep{56} to infer pixel-wise grasp quality, which achieved faster grasping. Recently, Cao et al. used an auto-encoder--decoder to infer the grasp quality, which was labeled by a similar contact model to that used in GQ-CNN, to generate the suction pose. However, the accuracy of the contact model depends on the model complexity and parameter tuning. High complexity may lead to long computation cost of annotation. Parameter identification by real world experiment \citep{57} might also be necessary to ensure the validity of the contact model. 

Our approach also labeled the grasp quality in synthesized depth images. Unlike GQ-CNN, we proposed a more intuitive evaluation metric based on a geometrical analytic method rather than a complicated contact analytic model. Our results showed that the intuitive evaluation metric was competitive with GQ-CNN. A reachability heatmap was further incorporated to help filter pixels that had high grasp quality value but were unreachable.

\subsection{Reachability assessment}

Reachability was previously assessed by sampling a large number of grasp poses and then using forward kinematics calculation, inverse kinematics calculation, or manipulability ellipsoid evaluation to investigate whether the sampled poses were reachable \citep{31,32,33,34,35}. The reachability map was generated off-line, and the feasibility of candidate grasp poses were queried during grasp planning for picking static \citep{36, 37} or moving \citep{38} objects. However, creating an off-line map with high accuracy for a large space is computationally expensive. In addition, although the off-line map considered only collisions between the manipulator and a constrained environment (e.g., fixed bin or wall), since the environment for picking in a cluttered scene is dynamic, collision checking between the manipulator and surrounding objects is still needed and this can be time consuming. Hence, recent studies have started to learn reachability with collision awareness of grasp poses. Kim et al. \citep{42} designed a density net to learn the reachability density of a given pose, but considered only self-collision. Murali et al. \citep{41} used a learned grasp sampler to sample 6D grasp poses and proposed a CollisionNet to assess the collision score of sampled poses. Lou et al. \citep{39} proposed a 3D CNN and reachability predictor to predict the pose stability and reachability of sampled grasp poses. They later extended the work by incorporating collision awareness for learning approachable grasp poses \citep{40}. These sampling-based methods have required designing or training a good grasp sampler for inferring the reachability. Our approach is one-shot, which directly infers the pixel-wise reachability from the depth image without sampling.  

\section{Problem statement}
\subsection{Objective}

Based on depth image and point cloud input, the goal is to find a grasp pose with high graspability for a suction robotic hand to pick items in a cluttered scene and then place them on a conveyor. The depth image and point cloud point are directly obtained from an Intel RealSense SR300 camera.

\subsection{Picking robot}

\begin{figure}[h!]
\begin{center}
\includegraphics[scale=0.7]{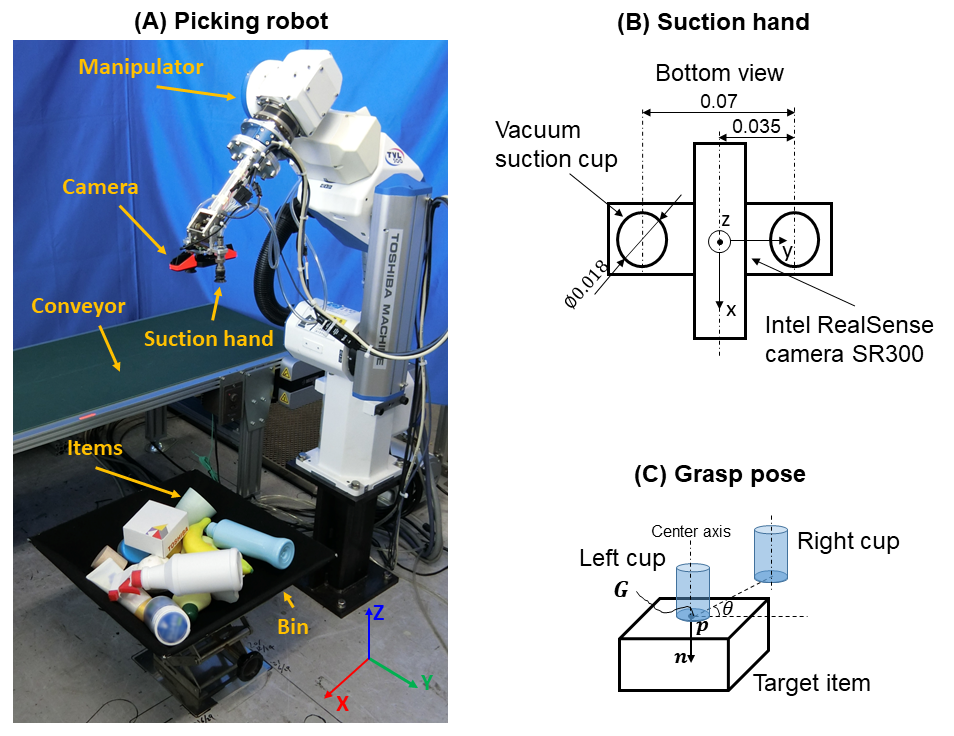}
\end{center}
\caption{Problem statement: \textbf{(A)} Picking robot; \textbf{(B)} Suction hand; \textbf{(C)} Grasp pose.}\label{fig:1}
\end{figure}

As shown in Fig. \ref{fig:1} (A), the picking robot is composed of a 6 degree-of-freedom (DoF) manipulator (TVL500, Shibaura Machine Co., Ltd.) and a 1 DoF robotic hand with two vacuum suction cups (Fig. \ref{fig:1} (B)). The camera is mounted in the center of the hand and is activated only when the robot is at its home position (initial pose), and hence can be regarded as a fixed camera installed above the bin. This setup has the merit that the camera can capture the scene of the entire bin from the view above the bin center without occlusion by the manipulator.

\subsection{Grasp pose}

As shown in Fig. \ref{fig:1} (C), the 6D grasp pose $\bm{G}$ is defined as ($\bm{p}, \bm{n},  \theta$), where $\bm{p}$ is the target point position of the vacuum suction cup center, $\bm{n}$ is the suction direction, and $\theta$ is the rotation angle around $\bm{n}$. Given the point cloud of the target item and $\bm{p}$ position, the normal of $\bm{p}$ can be calculated simply by principal component analysis of a covariance matrix generated from neighbors of $\bm{p}$ using a point cloud library. $\bm{n}$ is the direction of the calculated normal of $\bm{p}$. As $\bm{n}$ determines only the direction of the center axis of the vacuum suction cup, a further rotation degree of freedom ($\theta$) is required to determine the 6D pose of the hand. Note that the two vacuum suction cups are symmetric with respect  to the hand center.

\section{Method}

The overall picking system diagram is shown in  Fig. \ref{fig:2}. Given a depth image captured at the robot home position, the auto-encoder SG-U-Net++ predicts the suction graspability maps, including a pixel-wise grasp quality map and a robot reachability map. The auto-encoder SG-U-Net++ is trained using a synthesized dataset generated by a physical simulator without any human-labeled data. Cluster analysis is performed on two maps to find areas with graspbility higher than the thresholds. Local sorting is performed to extract the points with highest graspbility values in each cluster as grasp candidates. Global sorting is further performed to sort the candidates of all clusters in descending order of graspbility value, and this is sent to the motion planner. The motion planner plans the trajectory for reaching the sorted grasp candidates in descending order of graspability value. The path search continues until the first successful solution of the candidate is found. 

\subsection{Learning the suction graspability}

SG-U-Net++ was trained on a synthesized dataset to learn suction graspability by supervised deep learning. Figure \ref{fig:3} (A) shows the overall dataset generation flow. A synthesized cluttered scene is firstly generated using pybullet to obtain a systematized depth image and object segmentation mask. Region growing is then performed on the point cloud to detect the graspable surfaces. A convolution-based method is further used to find the graspable areas of vacuum cup centers where the vacuum cup can make full contact with the surfaces. The grasp quality and robot reachability are then pixel-wise evaluated by the proposed metrics in the graspable area. 

\subsubsection{Cluttered scene generation}

The object set used to synthesize the scene contains 3D CAD models from the 3DNet \citep{17} and KIT Object database \citep{18}. These models were used because they had previously been used to generate a dataset for which a trained CNN successfully predicted the grasp quality \citep{19}. We empirically removed objects that are obviously difficult for suction to finally obtain 708 models. To generate cluttered scenes, a random number of objects were selected from the object set randomly, and were dropped from above the bin in random poses. Once the state of all dropped objects was stable, a depth image and segmentation mask for the cluttered scene was generated, as in Fig. \ref{fig:3} (A).

\subsubsection{Graspable surface detection}

As shown in Fig. \ref{fig:3} (C), in order to find the graspable area of each object, graspable surface detection was performed. Given the camera intrinsic matrix, the point cloud of each object can be easily created from the depth image and segmentation mask. To detect surfaces that are roughly flat and large enough for suction by the vacuum cup, a region growing algorithm \citep{20} was used to segment the point cloud. To stably suck an object, the vacuum cup needs to be in full contact with the surface. Hence, inspired by \citep{11}, a convolution based method was used to calculate the graspable area (set of vacuum cup center positions where the cup could make full contact with the surface). Specifically, as shown in the middle of Fig. \ref{fig:3} (C), each segmented point cloud was projected onto its local coordinates to create a binary surface mask. Each pixel of the mask represents 1 mm. The surface mask was then convolved with a vacuum cup mask (of size ${18\times18}$, where 18 is the cup diameter) to obtain the graspable area. At a given pixel, the convolution result is the area of the cup ($\pi*0.009^2$ for our hand configuration) if the vacuum cup can make full contact with the surface. Refer to \citep{11} for more details. The calculated areas were finally remapped to a depth image to generate a graspable area map (right side of Fig. \ref{fig:3} (C)).

\subsubsection{Grasp quality evaluation}

Although the grasp areas of the surfaces were obtained, each pixel in the area may have different grasp probability, that is, grasp quality, owing to surface features. Therefore, an intuitive metric $J_q$ (Eq. (\ref{eq1})) was proposed to assess the grasp quality for each pixel in the graspable area. The metric $J_q$ is made up of $J_c$ which evaluates the normalized distance to the center of the graspable area and $J_s$ which evaluates the flatness and smoothness of the contact area between the vacuum cup and surface. 

\begin{equation}
J_{q} = 0.5J_{c}+0.5J_{s}
\label{eq1}
\end{equation}

$J_c$ (Eqs. (\ref{eq2})-(\ref{eq3})) was derived based on the assumption that the closer the grasp points are to the center of graspable area, the closer they are to the center of mass of the object. Hence, grasp points close to the area center (higher $J_c$ values) are considered to be more stable for the robot to suck and hold the object.  

\begin{equation}
J_{c} = 1-\textrm{maxmin}(\|\bm{p}-\bm{p}_c\|_2)
\label{eq2}
\end{equation}

\begin{equation}
\textrm{maxmin}(\bm{x}) = \frac{\bm{x}-\textrm{min}(\bm{x})}{\textrm{max}(\bm{x})-\textrm{min}(\bm{x})} 
\label{eq3}
\end{equation}
where $\bm{p}$ is a point in a graspable area of a  surface, $\bm{p}_c$ is the center of the graspable area, and $\textrm{maxmin}(\bm{x})$ is a max-min normalization function.

$J_s$ (Eqs. (\ref{eq4})-(\ref{eq6})) was derived based on the assumption that a vacuum cup generates a higher suction force when in contact with a flat and smooth surface than a curved one. We defined $\bm{p}_s$ as the point set of the contact area between the vacuum cup and the surface when the vacuum cup sucked at a certain point in the graspable area. As reported in \citep{21}, the surface flatness can be evaluated by the variance of the normals, the first term of $J_s$ assesses the surface flatness by evaluating the variance of the normals of $\bm{p}_s$ as in Eq. (\ref{eq5}). However, it is not sufficient to consider only the flatness. For example, although a vicinal surface has small normal variance, the vacuum cup cannot achieve suction to this kind of step-like surface. Hence, the second term (Eq. (\ref{eq6})) was incorporated to assess the surface smoothness by evaluating the residual error to fit $\bm{p}_s$ to a plane $\textrm{Plane}(\bm{p_s})$ where sum of the distance of each point in $\bm{p}_s$ to the fitted plane is calculated. Note that we empirically scaled ${\textrm{res}(\bm{p_s})}$ by 5.0 and added weights 0.9 and 0.1 to two terms in Eq. \ref{eq4}  to obtain plausible grasp quality values.

\begin{equation}
J_{s} = 0.9\textrm{var}(\bm{n_s}) + 0.1e^{-5\textrm{res}(\bm{p_s})}
\label{eq4}
\end{equation}
\begin{equation}
\textrm{var}(\bm{n}_{s}) = \frac{\displaystyle\sum_{i=1}^{N}\bm{n}_{s,i}-\bar{\bm{n}}_s}{N-1} 
\label{eq5}
\end{equation}
\begin{equation}
\textrm{res}(\bm{p_s}) = \displaystyle\sum_{i=1}^{N}\|\bm{p}_{s,i}-{\textrm{Plane}(\bm{p_s})}\|_2
\label{eq6}
\end{equation}
where $\bm{p_s}$ are the points in the contact surface when the vacuum cup sucks at a point in the graspable area, $N$ is the number of points in $\bm{p_s}$, $\bm{n_s}$ are the point normals of $\bm{p_s}$, $\textrm{var}(\bm{n}_{s})$ is the function to calculate the variance of $\bm{n}_{s}$, $\textrm{Plane}(\bm{p_s})$ is a plane equation fitted by $\bm{p_s}$ using the least squares method, and $\textrm{res}(\bm{p_s})$ is the function to calculate the residual error of the plane fitting by calculating the sum of distance from each point in $\bm{p_s}$ to the fitted plane.

Figure \ref{fig:3} (D) shows an example of the annotated grasp quality. Points closer to the surface center had higher grasp quality values, and points located on flat surfaces had higher grasp quality (e.g., surfaces of boxes had higher grasp quality values than cylinder lateral surfaces). 

\subsubsection{Robot reachability evaluation}

The grasp quality considers only the interaction between the object and the vacuum cup without considering the manipulator. As a collision check and IK solution search for the manipulator are needed, on-line checking and searching for all grasp candidates is costly. Learning robot reachability helped to rapidly avoid the grasp points where the hand and manipulator may collide with the surroundings. It also assessed the ease of finding inverse kinematics (IK) solutions for the manipulator. 

As described in section 3.3, $\bm{p}$ and $\bm{n}$ of a grasp pose $\bm{G}$ can be calculated from the point cloud. $\theta$ is the only undetermined variable for defining a $\bm{G}$. We sampled the $\theta$ from $0^\circ$ to $355^\circ$ in step intervals of $5^\circ$. IKfast \citep{26} and Flexible Collision Library (FCL) \citep{27} were used to calculate the inverse kinematics solution and detect the collision check for each sampled $\theta$. The reachability evaluation metric (Eqs. (\ref{eq7})-(\ref{eq8})) assessed the ratio of the number of IK valid $\theta$ (had collision free IK solution) to the sampled size $N_{\theta}$. 

\begin{equation}
J_{a} = \frac{\displaystyle\sum_{i=1}^{N_{\theta}}\textrm{Solver}(\bm{p},\bm{n},\theta_i)}{N_{\theta}}
\label{eq7}
\end{equation}

\begin{equation}
\textrm{Solver}(\bm{p},\bm{n},\theta_i)=  \begin{cases}
                                   1 & \text{if collision free and IK solution exists} \\
                                   0 & \text{else} 
  \end{cases}
\label{eq8}
\end{equation}
where $N_{\theta}$ is the size of sampled $\theta$ and $\textrm{Solver}$ is the IK solver and collision checker for the robot.

Note that because the two vacuum cups are symmetric with respect to the hand center, we evaluated the reachability score of only one cup. Figure \ref{fig:3} (E) shows an example of the robot reachability evaluation.

\subsubsection{SG-U-Net++}

As shown in Fig. \ref{fig:4}, a nest structured auto-encoder--decoder called suction graspability U-Net++ (SG-U-Net++) was used to learn the suction graspability. We used the nested architecture because it was previously reported to have high performances for semantic segmentation.  Given a $256\times256$ depth image, SG-U-Net++ outputs $256\times256$ shape grasp quality and robot reachability maps. SG-U-Net++ resembles the structure of U-Net++ proposed by \citep{22}. SG-U-Net++ consists of several sub encoder--decoders connected by skip connections. For example, $X^{0,0} \rightarrow X^{1,0} \rightarrow X^{0,1}$ is one of the smallest sub encoder--decoders, and  $X^{0,0} \rightarrow X^{1,0} \rightarrow X^{2,3}\rightarrow X^{3,0} \rightarrow X^{4,0}\rightarrow X^{3,1} \rightarrow X^{2,2}\rightarrow X^{1,3} \rightarrow X^{0,4}$ is the largest encoder--decoder. The dense block for $X^{i,j}$ consists of two $3\times3\times32*2^i$ convolution (conv) layers, each of which is followed by batch normalization and ReLu activation. The output layer connected to $X^{0,4}$ is a $1\times1\times2$ conv layer. MSELoss was used for supervised pixel-wise heatmap learning.

\subsection{Clustering and ranking}

\begin{figure}[h!]
\begin{center}
\includegraphics[scale=0.25]{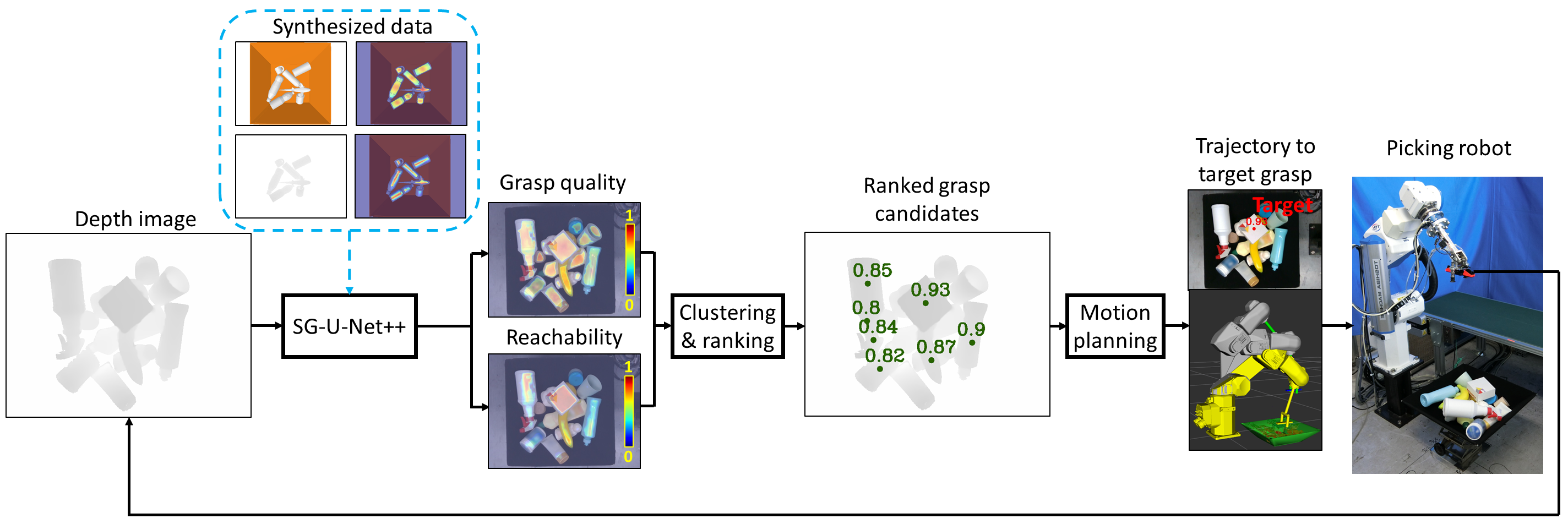}
\end{center}
\caption{System diagram.}\label{fig:2}
\end{figure}

\begin{figure}[h!]
\begin{center}
\includegraphics[scale=0.4]{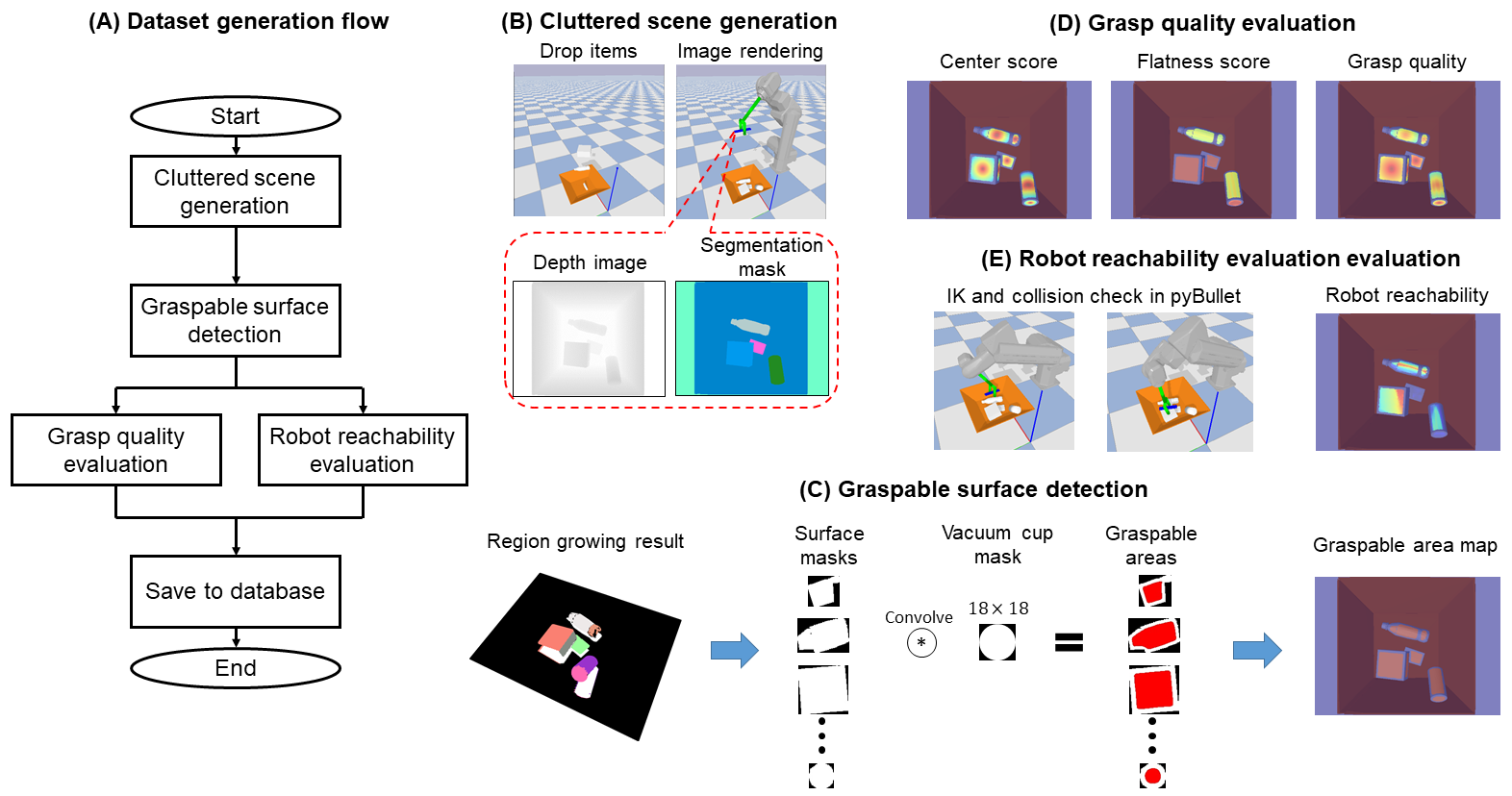}
\end{center}
\caption{Data generation pipeline: \textbf{(A)} Dataset generation flow; \textbf{(B)} Cluttered scene generation; \textbf{(C)} Graspable surface detection; \textbf{(D)} Grasp quality evaluation; \textbf{(E)} Robot reachability evaluation.}\label{fig:3}
\end{figure}

\begin{figure}[h!]
\begin{center}
\includegraphics[scale=0.4]{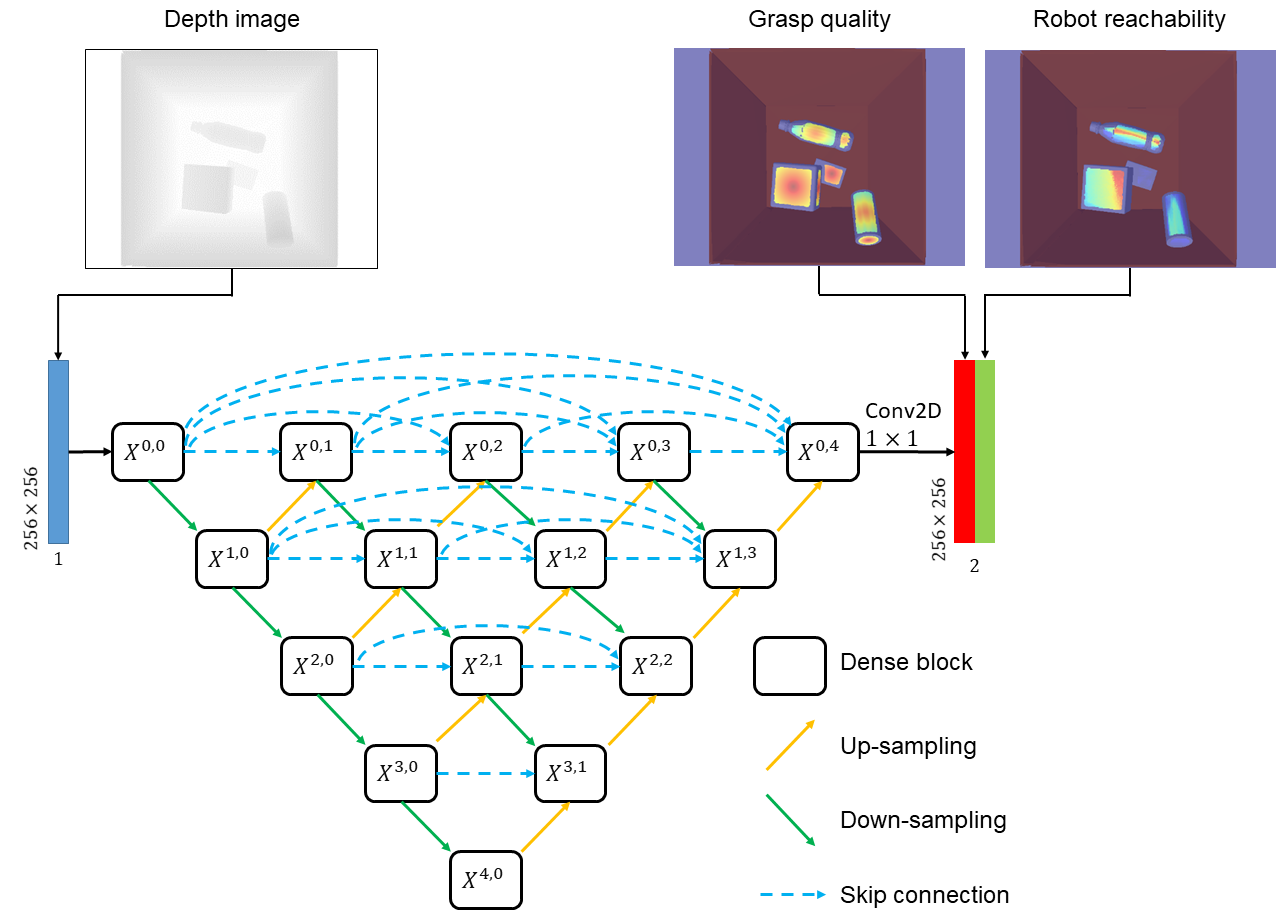}
\end{center}
\caption{Architecture of SG-U-Net++.}\label{fig:4}
\end{figure}

The clustering and ranking block in Fig. \ref{fig:2} outputs the ranked grasp proposals. To validate the role of learning  reachability, we proposed two policies (Policy 1: use only grasp quality; Policy 2: use both grasp quality and reachability) to propose the grasp candidates. Policy 1 extracted the area of grasp quality values larger than threshold $th_g$. Policy 2 extracted the area of grasp quality score values larger than threshold $th_g$ and the corresponding reachability score values larger than $th_r$. Filtering by reachability score value was assumed to help to remove pixels with high grasp quality value that are not reachable by the robot due to collision or IK error. The values of $th_g$ and $th_r$ were empirically set to 0.5 and 0.3, respectively. The extracted areas were clustered by scipy.ndimage.label \citep{28}. Points in each cluster were ranked (local cluster level) by the grasp quality values, and the point with the highest grasp quality was used as the grasp candidate for its owner clusters (see Ranked grasp candidates in Fig. \ref{fig:2}). Finally, the grasp candidates were further ranked (global level) and sent to the motion planner.

\subsection{Motion planning}

Given the grasp candidates, goal poses were created for moveIt \citep{29} to plan a trajectory. As described in 3.3, the values of $\bm{p}$ and $\bm{n}$ of a goal pose could be obtained from the corresponding point cloud information of the grasp candidates so that only $\theta$ was undetermined. As a cartesian movement path is required for the hand to suck the object, $\bm{p}$ was set to a 1 cm offset away from the object along the $\bm{n}$ direction. $\theta$ was sampled from $0^\circ$ to $180^\circ$ at step intervals of $5^\circ$. For each sampled goal pose, the trajectory was planned for left and right vacuum cup respectively, and the shorter trajectory was selected as the final solution. The planned trajectory was further time parametrized by Time-Optimal Path Parameterization (toppra) \citep{30} to realize position control for the robot to approach the goal pose. After reaching the goal pose, the robot hand moved down along $\bm{n}$ to suck the object. Once the contact force between the vacuum cup and object, which was measured by a force sensor, exceeded the threshold, the object was assumed to be sucked by the vacuum cup and was then lifted and placed on the conveyor.

\section{Experiments}
\subsection{Data collection, training, and precision evaluation}
We used the proposed suction graspability annotation method in pyBullet to generate 15,000 data items, which were split into 10,000 for training and 5,000 for testing. The synthesized data was then used to train SG-U-Net++, which was implemented by pyTorch. The adam optimizer (learning rate = 1.0e-4) was used to update the parameters of the neural network during the training. The batch size was set to 16. Both data collection and training were conducted on an Intel Core i7-8700K 3.70 GHz PC with 64G RAM and 4 Nvidia Geforce GTX 1080 GPUs.

To evaluate the learning results, we used a similar evaluation method to that reported in Zeng et al. \citep{14} on the testing set. For practical utilization, it is important for SG-U-Net++ to find at least one point in ground truth suction graspable area or manipulator reachable area. We defined suction graspable area as the pixels whose ground truth grasp quality scores are larger than 0.5, and approachable area as the pixels whose ground truth reachability scores are larger than 0.5. The inferred grasp quality and reachability scores were divided by thresholds into Top 1\%, Top 10\%, Top 25\%, and Top 50\%. If pixels larger than the threshold were located in the ground truth area, the pixels were considered true positive, otherwise the pixels were considered false positive. We report the inference precision for the four thresholds above for SG-U-Net++, and compared them with Dex-Net.

\subsection{Real world picking experiments}

\begin{figure}[h!]
\begin{center}
\includegraphics[scale=0.6]{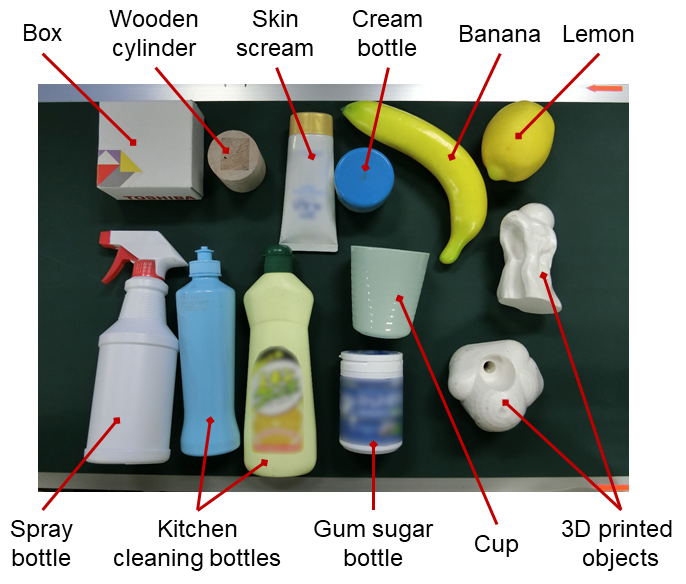}
\end{center}
\caption{Experiment object set.}\label{fig:5}
\end{figure}

To evaluate and compare the performance of different policies for the picking system, a pick-and-placement task experiment was conducted. Figure \ref{fig:5} shows the experimental object set, which includes primitive solids, commodities, and 3D-printed objects. All of the objects are novel objects that were not used during training. During each trial, the robot was required to pick 13 randomly posed objects (except for the cup) from a bin and then place them on the conveyor.  Note that the cup was placed in the lying pose because it could not be grasped if it was in a standing pose. A grasp attempt was treated as a failure if the robot could not grasp the object in three attempts.

We conducted 10 trials for Policy 1, Policy 2, and Dex-Net 4.0 (suction grasp proposal by fully convolutional grasping policy), respectively. Note that because Dex-Net had its own grasp planning method, we directly sorted the inferred grasp quality values without clustering. To compare our proposed intuitive grasp quality evaluation metric (Eq. (\ref{eq1})) with the one used in Dex-Net, we evaluated and compared the grasp planning computation time cost and success rate of Policy 1 and Dex-Net. To evaluate the effect of incorporating the reachability score, we evaluated and compared the grasp planning computation time cost, motion planning computation time cost, and success rate of Policy 1 and Policy 2.

\section{Results and discussion}
\subsection{Inference precision evaluation}

Table \ref{tab:table1} shows the inference precision of grasp quality and reachability. Both SQ-U-Net++ and Dex-Net achieved high precisions for Top 1\% and Top 10\%, but the precision of Dex-Net decreased to lower than 0.9 for Top 25\% and Top 50\%. This result indicates that our proposed intuitive grasp quality evaluation metric (Eq. (\ref{eq1})) performed as well as a physically inspired evaluation metric. Learning the suction graspability annotation by point cloud analytic methods might not be so bad compared to dynamics analytic methods for the suction grasp task. However, the inference precision of the reachability for SQ-U-Net++ also achieved larger than 0.9 for Top 1\% and Top 10\%, but decreased sharply for Top 25\% and Top 50\%. The overall performance of reachability inference was poorer than grasp quality, indicating that reachability is more difficult to learn than grasp quality. This is probably because grasp quality can be learned from the surface features, but reachability learning requires more features such as the features of surrounding objects in addition to the surface features, leading to more difficult learning.

\subsection{Picking experiments}
\subsubsection{Overall performance}

Table \ref{tab:table2} shows the experimental results of Dex-Net and our proposed method. Although all three methods achieved a high grasp success rate ($>90\%$), our method took shorter time for grasp planning. Moreover, the motion planning computation time was reduced by incorporating the learning of reachability. The SQ-U-Net++ Policy 2 achieved a high speed picking of approximately 560 PPH (piece per hour) (see supplementary video).

\subsubsection{Comparison with physically-inspired grasp quality evaluation metric}

\begin{figure}[h!]
\begin{center}
\includegraphics[scale=0.5]{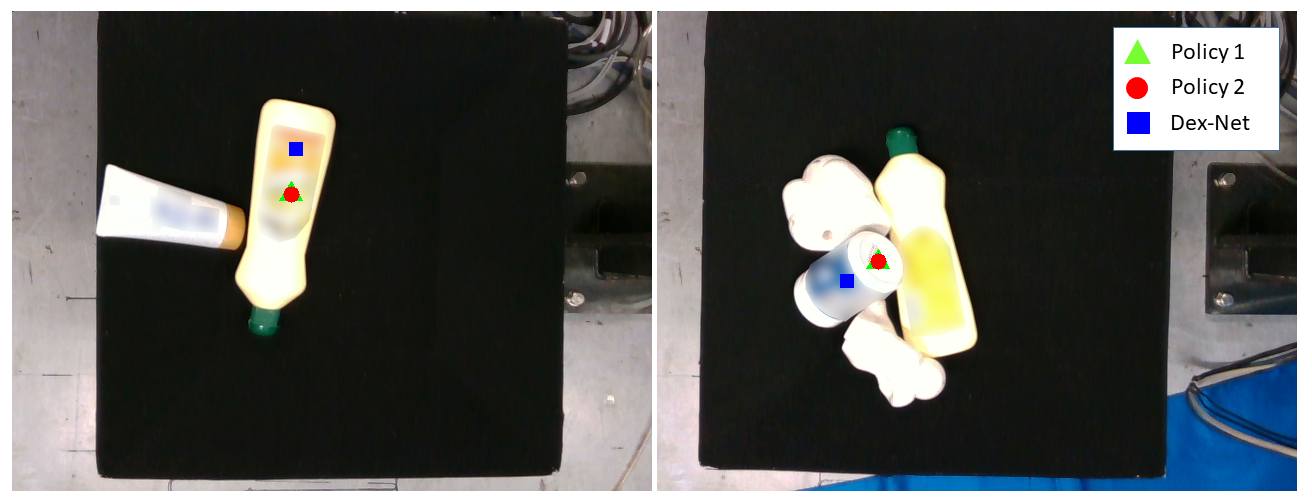}
\end{center}
\caption{Example of Dex-Net grasp prediction that is farther from the center of mass of the object.}\label{fig:6}
\end{figure}

As shown in Table \ref{tab:table2}, although our method was competitive with Dex-Net, it was faster for grasp planning. This result indicates that our geometric analytic based grasp quality evaluation is good enough for the picking task compared with a physically-inspired one. The evaluation of contact dynamics between a vacuum cup and the object surface might be simplified to just analyze the geometric features of the vacuum cup (e.g., shape of the cup) and surfaces (e.g., surface curvature, surface smoothness, and distance from the cup center to the surface center). In addition, similar to the report in \citep{14}, the grasp proposal of Dex-Net was farther from the center of mass. Figure \ref{fig:6} shows an example of our method and Dex-Net. Our predicted grasps were closer to the center of mass of the object than the ones inferred by Dex-Net. This is because we incorporated $J_c$ (Eq. \ref{eq2}) to evaluate the distance from the vacuum cup center to the surface center, helping the SQ-U-Net++ to predict grasp positions much closer to the center of mass.

\subsubsection{Role of learning reachability}

\begin{figure}[h!]
\begin{center}
\includegraphics[scale=0.5]{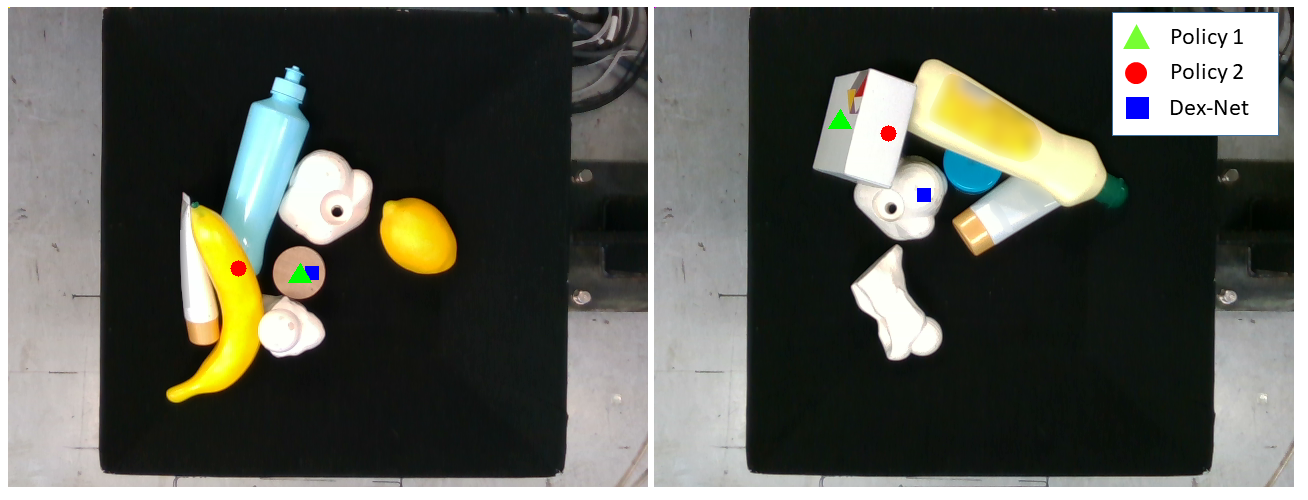}
\end{center}
\caption{Example of grasps predicted by Dex-Net and Policy 1 that are unreachable or difficult to reach.}\label{fig:7}
\end{figure}

Although learning reachability increased the grasp planning cost a little bit by 0.02 s due to the processes such as clustering and ranking of the reachability heatmap, it helped to reduce the motion planning cost (Policy 2: 0.90 s vs. Policy 1: 1.71s). As show in Fig. \ref{fig:7}, Policy 2 predicted grasps with lower collision risks with neighboring objects than did Policy 1 and Dex-Net (e.g., Fig. 7 Left: Policy 1 and Dex-Net predicted grasps on wooden cylinder that had high collision risks between the hand and 3D printed objects). Further, an object might have surfaces with the same grasp quality (e.g., Fig. 7 Right: box with two flat surfaces). Whereas Policy 2 selected the surface that was easier to reach, Policy 1 might select the one that is difficult to reach (Fig. 7 Right), since it does not consider the reachability. Therefore, Policy 2 was superior to Policy 1 and Dex-Net because it removed the grasp candidates that were obviously unable or difficult to approach. However, for Policy 1 and Dex-Net, as they considered only the grasp quality, the motion planner might firstly search the solutions for the candidates with high grasp quality, but those candidates might be unreachable for the manipulator, and thus increase the motion planning effort.

\subsubsection{Limitations and future work}
Our study was not devoid of limitations. Several grasp failures occurred when picking 3D printed objects. Since the synthesized depth images differ from real ones because real images are noisy and incomplete, the neural network prediction error increased for real input depth images. This error was tolerable for objects with larger surfaces like cylinders and boxes, but intolerable for 3D printed objects that have complicated shapes where the graspable areas are quite small. In the future, we intend to conduct sim-to-real \citep{58} or depth missing value prediction \citep{59} to improve the performance of our neural network. Another failure was that although not very often, the objects fell down during holding and placement because the speed of the manipulator was too high to hold the object stably. We addressed this problem by slowing down the manipulator movement during the placement action but this sacrificed the overall system picking efficiency. In the future, we want to consider a more suitable method for object holding and placement trajectory such as model based control.

\section{Conclusion}

We proposed an auto-encoder--decoder to infer the pixel-wise grasp quality and reachability. Our method is intuitive but competitive with CNN trained by data annotated using physically-inspired models. The reachability learning improved the efficiency of the picking system by reducing the motion planning effort. However, the performance of the auto-encoder--decoder deteriorated because of differences between synthesized and real data. In the future, sim-to-real technology will be adopted to improve the performance under various environments.  

\section*{Data Availability Statement}
The original contributions presented in the study are included in the article/supplementary material, further inquiries can be directed to the corresponding author.

\section*{Author Contributions}

PJ made substantial contributions to conceiving the original ideas, designing the experiments, analyzing the results, and writing the original draft. JO, YI, JO, HH, AS, ST, HE, KK, and AO helped to conceptualize the final idea. PJ, YI, and JO conducted the experiments. PJ, YI, and JO revised the manuscript. YI and AO supervised the project. All authors contributed to the article and approved the submitted version.

\section*{Funding}
The authors received no financial support for the research, authorship, and/or publication of this article.

\section*{Conflict of Interest Statement}

The authors declare that the research was conducted in the absence of any commercial or financial relationships that could be construed as a potential conflict of interest.


\bibliographystyle{unsrtnat}
\bibliography{references}  






\newpage

\begin{table}[h!]
    \caption{Inference precision.}
    \label{tab:table1}
    \begin{tabular}{c|c|cccc}
    \hline
      Score & Method & Top 1\% & Top 10\%& Top 25\%& Top 50\%\\ \hline
      \multirow{ 2}{*}{Grasp quality} &Dex-Net & 91.9 & 91.0 & 88.7 & 84.2\\
      &SQ-U-Net++ & 99.8 & 99.6 & 99.2 & 97.5\\
    \hline
    Reachability & SQ-U-Net++ & 95.8 & 91.1 & 80.7 & 61.2\\
    \hline
    \end{tabular}
\end{table}

\begin{table}[h!]
    \caption{Experiment results.}
    \label{tab:table2}
    \begin{tabular}{c|c|c|c}
    \hline
      Method & Success rate (\%) & Grasp planning cost (s) & Motion planning cost (s)\\ \hline
      Dex-Net 4.0 Suction & \multirow{ 2}{*}{91.5} & \multirow{ 2}{*}{0.60} & \multirow{ 2}{*}{2.91}\\
      (FC-GQCNN-4.0-SUCTION)&&&\\
      SQ-U-Net++ Policy1 & \multirow{ 2}{*}{94.6} & \multirow{ 2}{*}{$\bm{0.15}$} & \multirow{ 2}{*}{1.71}\\
      (grasp quality only)&&&\\
      SQ-U-Net++ Policy2 & \multirow{ 2}{*}{$\bm{95.4}$} & \multirow{ 2}{*}{0.17} & \multirow{ 2}{*}{$\bm{0.90}$}\\
      (grasp quality + reachability)&&&\\
    \hline
    \end{tabular}
\end{table}

\end{document}